\documentclass{article}

\usepackage{arxiv}

\usepackage[utf8]{inputenc} 
\usepackage[T1]{fontenc}    
\usepackage{hyperref}       
\usepackage{url}            
\usepackage{booktabs}       
\usepackage{amsfonts}       
\usepackage{nicefrac}       
\usepackage{microtype}      
\usepackage{lipsum}
\usepackage{graphicx}
\graphicspath{ {./images/} }
\usepackage{adjustbox}
\usepackage{amsmath}

\title{Semantic-driven Colorization}

\author{
 Man M. Ho \\
  Hosei University, Japan\\
  \texttt{man.hominh.6m@stu.hosei.ac.jp}
   \And
 Lu Zhang \\
  Univ Rennes, INSA Rennes, CNRS\\
  IETR - UMR 6164, France\\
  \texttt{lu.ge@insa-rennes.fr}
  \And
 Alexander Raake\\
  Audiovisual Technology Group\\
  TU Ilmenau, Germany\\
  \texttt{alexander.raake@tu-ilmenau.de}
  \And
 Jinjia Zhou\\
  Hosei University, Japan\\
  \texttt{jinjia.zhou.35@hosei.ac.jp}
}

\begin{document}
\maketitle

\begin{figure}[htbp]
\centering
\includegraphics[width=\textwidth]{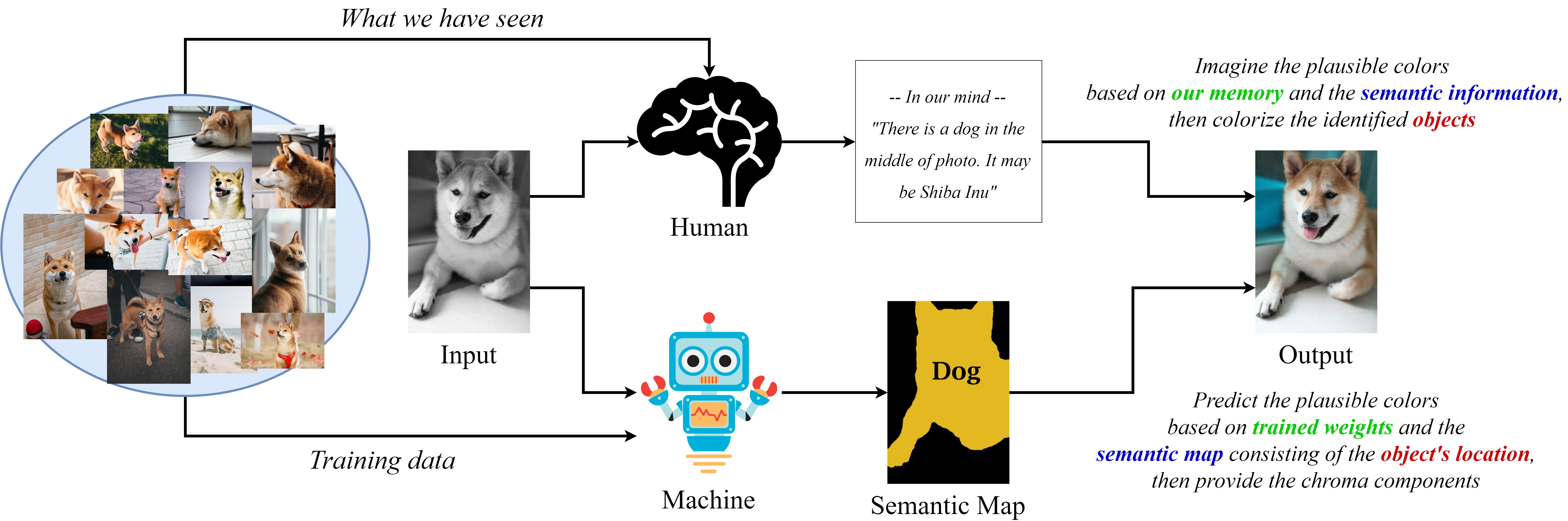}
\caption{We apply the human-like action in coloring a black-and-white photo to learning-based colorization.}
 \label{fig:teaser}
\centering
\end{figure}

\begin{abstract}
Recent colorization works implicitly predict the semantic information while learning to colorize black-and-white images. Consequently, the generated color is easier to be overflowed, and the semantic faults are invisible. As a human experience in colorization, our brains first detect and recognize the objects in the photo, then imagine their plausible colors based on many similar objects we have seen in real life, and finally colorize them, as described in Figure \ref{fig:teaser}. In this study, we simulate that human-like action to let our network first learn to understand the photo, then colorize it. Thus, our work can provide plausible colors at a semantic level. Plus, the semantic information of the learned model becomes understandable and able to interact. Additionally, we also prove that Instance Normalization is also a missing ingredient for colorization, then re-design the inference flow of U-Net to have two streams of data, providing an appropriate way of normalizing the feature maps from the black-and-white image and its semantic map. As a result, our network can provide plausible colors competitive to the typical colorization works for specific objects. Our interactive application is available at \url{https://github.com/minhmanho/semantic-driven_colorization}.
\end{abstract}


\section{Introduction}
Colorization is to generate colors for the old black-and-white photos. Thanks to the breakthrough of machine learning in many computer vision tasks, computers can handle the mentioned work surprisingly well. The common approaches are data-driven automatic colorization and user-guided colorization.

\begin{figure}[t]
    \centering
    \includegraphics[width=\textwidth]{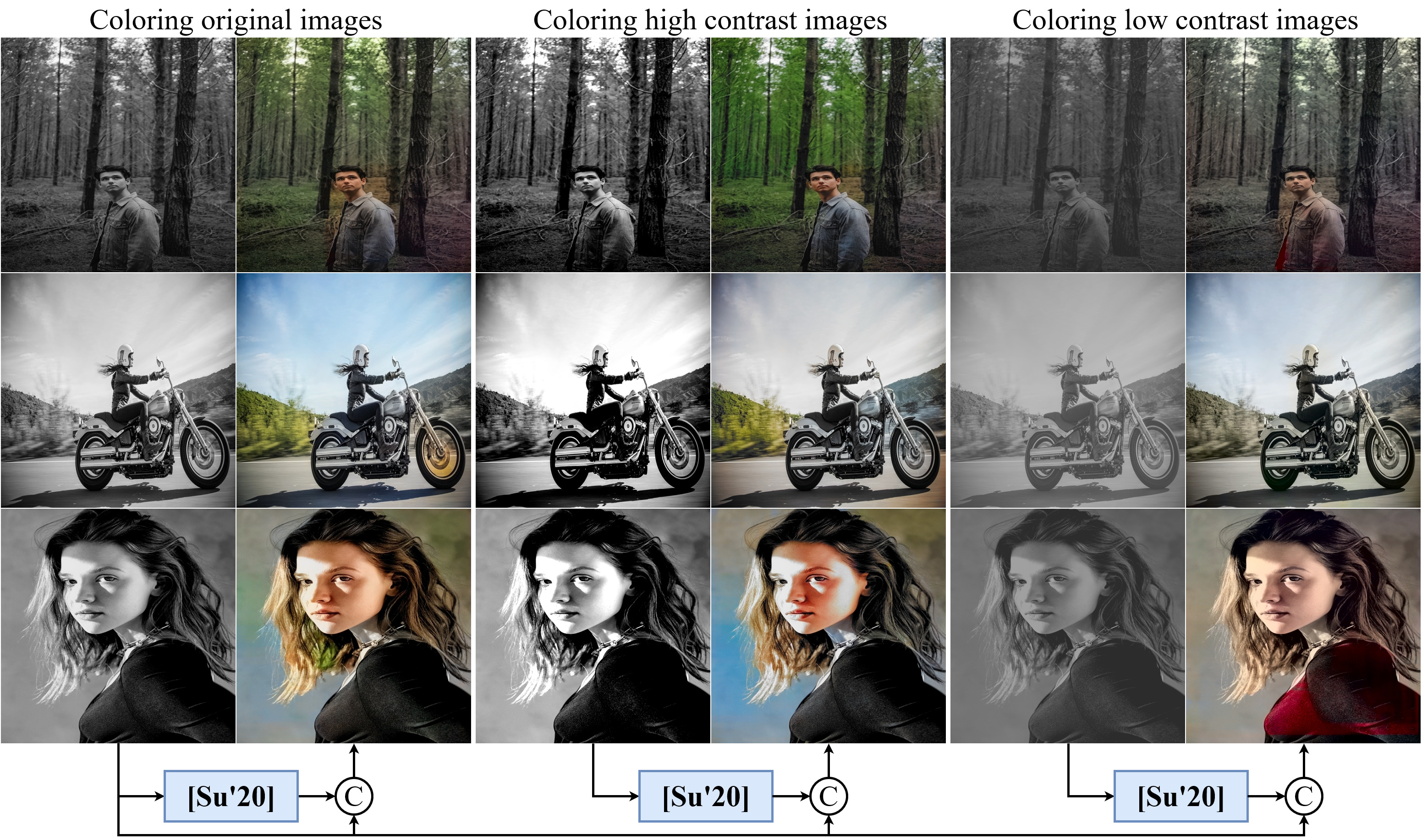}
    \caption{The contrast of the black-and-white images significantly affects the colors generated by Su'20 \cite{Su-CVPR-2020}, especially when the contrast is low. Therefore, Instance Normalization \cite{insnorm_ulyanov16} is also a missing ingredient in Image Colorization.}
    \label{fig:contrast}
\end{figure}

\begin{figure}[t]
\centering
\includegraphics[width=\textwidth]{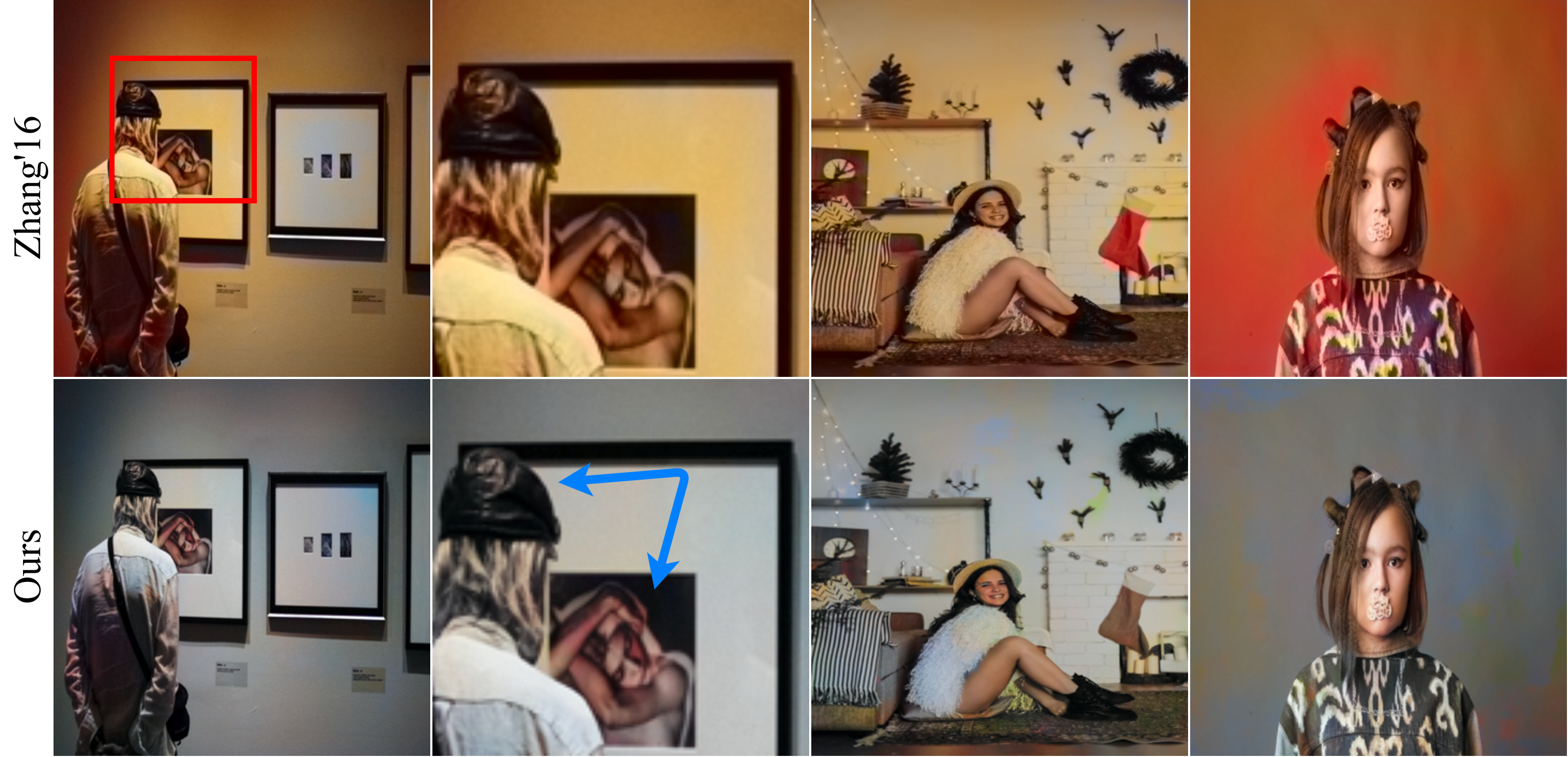}
\caption{The monochrome problem is that objects and backgrounds have varying tones of only one color in colorization. By simulating human-like action in coloring black-and-white photos, our work provides more distinct colors between objects compared with Zhang'16 \cite{zhang2016colorful}.}
\label{fig:monochrome}
\centering
\end{figure}

\textbf{Data-driven automatic techniques}. automatically generate plausible colors by learning a color mapping from the training data. Typically, an end-to-end deep neural network is employed to predict the colors of a black-and-white photo directly. Larsson et al. \cite{larsson2016learning} use the image classification pre-trained model of VGG16 \cite{simonyan2014very} to extract contextual information as a hyper-column, then leverage it to predict the Hue and Chroma components. Iizuka et al. \cite{iizuka2016let} divide their network into two streams for colorization and image classification. They resize the input image to a smaller size and feed it into the stream of image classification, then fuse the recognized features into the middle of the colorization stream. Their network is thus aware of the perceptual information of the input image. Meanwhile, Victoria et al. \cite{vitoria_2020_WACV} let the network learn image classification internally. Instead of the network being simultaneously trained with image classification, Yoo et al. \cite{yoo2019coloring} present a memory network to retrieve a color feature that best matches the ground-truth color feature, then modify the features in the middle of their colorization network using Adaptive Instance Normalization. Zhang et al. \cite{zhang2016colorful} solve the multimodal nature of the colorization problem by learning the color probabilities for their possible colors as a classification task. By adopting the legacy of image segmentation using deep learning, Zhao et al. \cite{zhao2018pixel} improve the work of Zhang et al. \cite{zhang2016colorful} to let the network also learn the meaning of pixels by detecting the semantic map while predicting colors. Afterward, they improve their colorization to learn the semantic segmentation inside their network, as described in \cite{zhao2019pixelated}. However, image classification and semantic segmentation are limited on gray-scale images, leading to semantic problems, such as color bleeding and color inconsistency. Moreover, semantic faults also bring monochrome-like problems. For example, Figure~\ref{fig:monochrome} shows the problem in the typical work \cite{zhang2016colorful} that \textit{objects} usually have the same color with the \textit{background}. As an experience in photo editing, the monochrome problem can be reproduced by step-by-step reducing Saturation (in Hue-Saturation-Lightness color space) and overlaying a color filter on an image. To solve this problem, Su et al. \cite{Su-CVPR-2020} proposed a way of focusing on coloring detected instances. However, they may suffer from low performance when none of the instances is detected, or the details other than "instances" are ignored. Plus, the instance-specific contrast information, which significantly affects their colorization performance, is not considered, as shown in Figure \ref{fig:contrast}. This work provides a scheme to predict colors based on the gray values and its semantic map at low-level features and designs a new inference flow for u-style architecture so that the instance-specific contrast information of gray values is properly removed. Therefore, our network can effectively learn the plausible colors for specific objects. Besides, the boundaries of objects in the black-and-white photo are identified before coloring, preventing color overflowing and bleeding, especially the monochrome problem. Therefore, our colorization network trained with semantic maps can provide the more distinct and plausible colors, as shown in Figure \ref{fig:monochrome}.

\textbf{User-guided edit propagation techniques}. rely on the user's suggestions to colorize the image. The user-guided input can be scribbles, color dots, or texts to control the generated colors. For example, Levin et al. \cite{levin2004colorization} use an optimization-based method to leverage the user's strokes to match it with the gray image. Meanwhile, Bahng et al. \cite{coloringwithwords2018} extract palette to guide their network based on user-provided words. Regarding exemplar-based colorization, the methods \cite{chia2011semantic, deshpande2017learning, gupta2012image, he2018deep, messaoud2018structural, welsh2002transferring} compute the color histogram of a reference image and transfer it to the gray-scale input. Instead of using another colorful image, Zhang et al. \cite{zhang2017real} present a way of using several user-guided color dots to colorize the input. Afterward, Yi Xiao et al. \cite{xiao2018interactive} adopt the work of Zhang et al. \cite{zhang2017real} to build a network that can support both a color histogram and user-guided color dots as global and local inputs simultaneously. Thanks to the great efforts of researchers in this field, the methods leveraging user-guided color dots achieve state-of-the-art performance. However, the given gray-scale pixel has various colors, especially those with the same light intensity. It is challenging for the user to avoid the incompatible colors between the objects. In fact, the harmony of a colorful image depends on the color of specific objects and the color between them. Most of the user-guided methods using color dots try to answer the question: "\textit{There are guided color dots, what should the color of this picture be?}" Meanwhile, this work tries to answer the question: "\textit{There is a dog on the grass under the sky, which color should this picture be?}" Thus, our method tends to synthesize the color based on not only gray-scale values but also the semantic information.

\begin{figure}[t]
\begin{center}
\includegraphics[width=\textwidth]{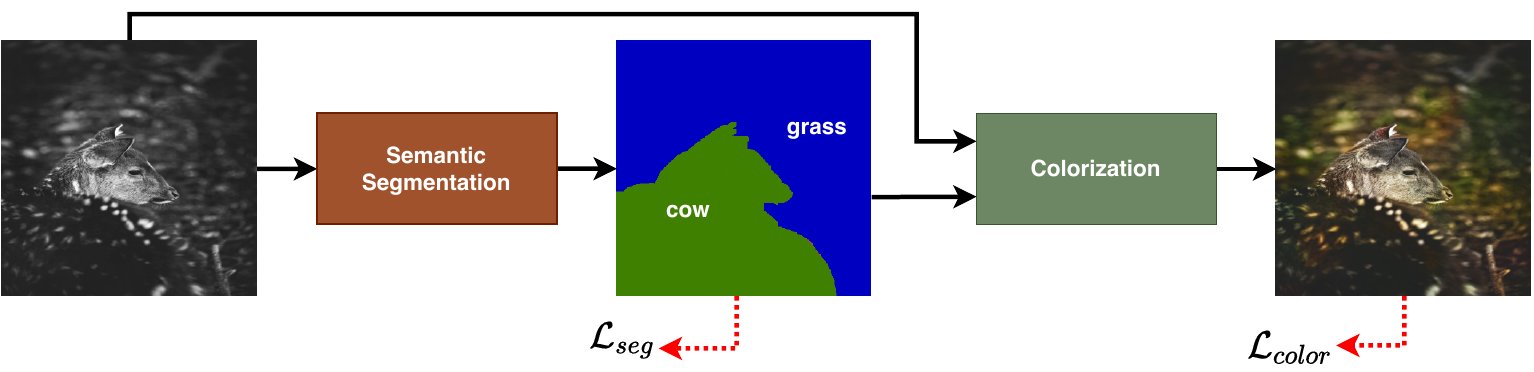}
\end{center}
\caption{The overview of the proposed system. We first predict a coarse semantic map revealing how our well-trained model sees the objects, then colorize the black-and-white input.}
\label{fig:concept}
\end{figure}

\section{The Proposed Colorization System}
In this study, we adopt the U-Net \cite{ronneberger2015u} and GridNet \cite{fourure2017residual} to design our colorization framework that can simulate human-like action, which is to first recognize the objects then colorize them, to enhance colorization performance. Our motivation is from (a) the generated colors can be more natural if we colorize gray-scale values based on its semantic information (e.g., which and how \textit{"red"} the color is to be plausible for the gray-scale values of \textit{"fire truck"}). While solving (a), we consider how to exploit the semantic information effectively as follows: (b) the object recognition task (e.g., image classification, semantic segmentation) is usually limited due to lack of color information and semantic faults are unable to be adjusted, (c) the instance-specific contrast information of the features extracted from the gray-scale image are complicated, significantly affecting on the generated colors, as shown in Figure \ref{fig:contrast}. To solve (a) and (b), we add a semantic segmentation network before the colorization network; therefore, semantic faults are visible and easily adjusted. However, the common way of concatenating many inputs cannot solve the issue (c) of learning from normalized gray-scale features and pure semantic features. We thus modify U-Net to have two streams of data: gray-scale image to extract gray features and segmentation map to extract semantic features. While inference, only gray features are simplified by an Instance Normalized (IN). An ablation study shows that training the network with simplified gray features can improve the colorization performance.

Our contributions are as follows:
\begin{itemize}
 \item As human-like action in colorizing a photo, we present a semantic-driven colorization framework that can detect objects, generalize the color based on the semantic information, and synthesize the colors competitive to the previous works. Furthermore, the detected semantic information can be visualized and adjusted.
 \item The contrast of a black-and-white image significantly affects the generated colors, revealing that Instance Normalization (IN) \cite{insnorm_ulyanov16} is also a missing ingredient for colorization. Therefore, gray features should be normalized. Instead of concatenating a black-and-white photo and its semantic map and having one stream of data, we modify the U-Net's inference stream to have two streams so that only gray-scale features are simplified. A qualitative result shows that the network with IN has a better performance.
 \item We build an interactive application to adjust the semantic information predicted by the learned segmentation model. Consequently, we can easily understand semantic-related issues of the whole colorization framework through the application and conveniently adjust the semantic map.
\end{itemize}

\begin{figure}[t]
\begin{center}
\includegraphics[width=\textwidth]{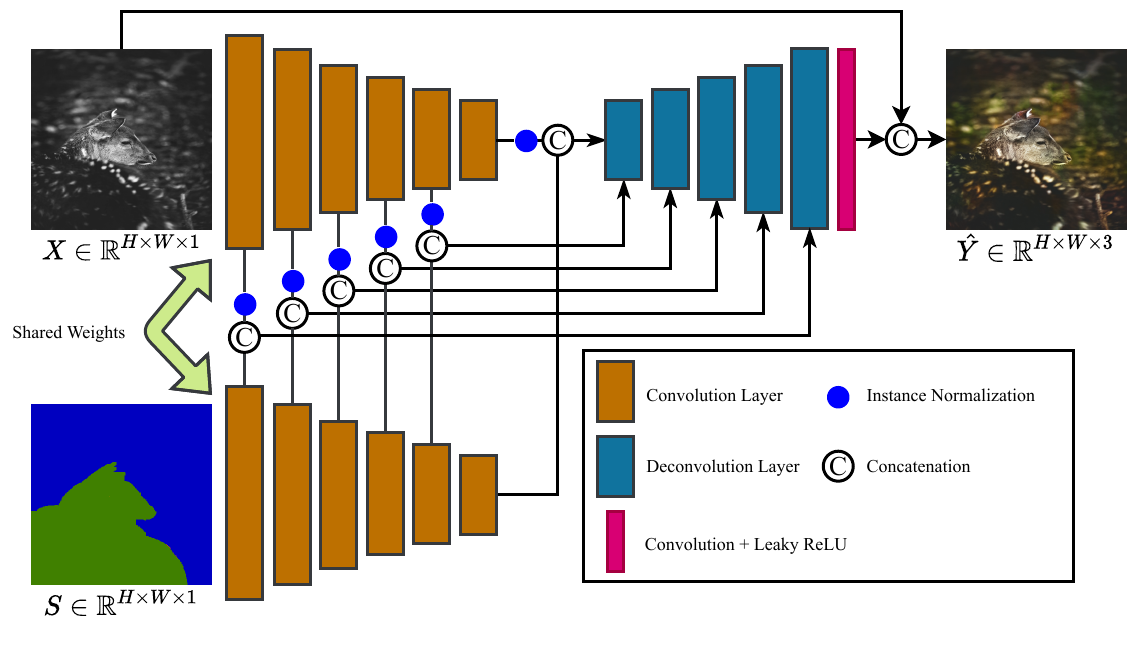}
\end{center}
\caption{U-Net with the modified inference flow for our Colorization. Features extracted from the gray-scale image are normalized with Instance Normalization before concatenation.}
\label{fig:color}
\end{figure}

\begin{figure}[t]
\centering
\includegraphics[width=0.8\textwidth]{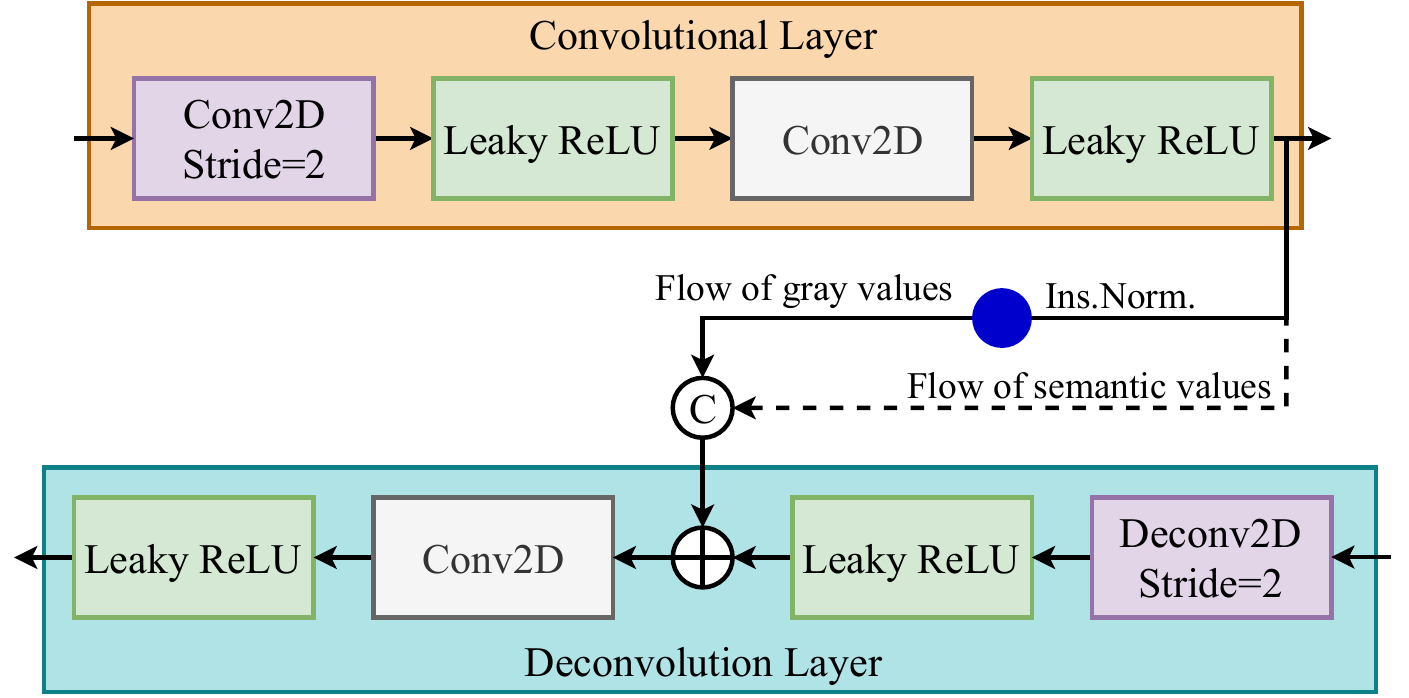}
\caption{Convolution Layer (CL) and Deconvolution Layer (DL) in Figure \ref{fig:color}. $\oplus$ and \textcircled{c} denote summation and concatenation, respectively.}
\label{fig:components}
\centering
\end{figure}

\subsection{System Overview}
As shown in Figure \ref{fig:concept}, the proposed colorization framework is composed of two components: Semantic Segmentation using GridNet \cite{fourure2017residual} and Colorization using the modified U-Net. Firstly, the gray image is fed to GridNet to predict the initial semantic map. Afterward, our colorization network will leverage the gray-scale image and its semantic map to generate the colors. Since the user can adjust the initial semantic map, we train our colorization network with the human-annotated semantic maps to make the colors more plausible. As mentioned, Semantic Segmentation and Colorization are trained independently.

Inspired by \cite{iizuka2016let,larsson2016learning,zhang2016colorful,zhang2017real}, we use CIE \textit{Lab} color space, which includes the luma and chroma representing gray-scale and color components, respectively. Given the luma $X \in \mathbb{R}^{H \times W \times 1}$ and the semantic map $S \in \mathbb{R}^{H \times W \times 1}$, our target is to generate chroma as the plausible color $\hat{Y} \in \mathbb{R}^{H \times W \times 2}$, which can be concatenated with $X$ to produce a colorful image. Training on the gray-scale values $X$ with its human-annotated semantic values $S$ will enhance the plausibility of generated colors when accurate semantic values are given in testing. Since completing the map $S$ is inconvenient for the user, we thus firstly use the re-trained GridNet \cite{fourure2017residual} to detect the coarse semantic map $\hat{S} \in \mathbb{R}^{H \times W \times 1}$. According to the predicted $\hat{S}$, the user can constantly adjust the map from coarse to fine to produce a satisfactory output. However, there is no user interaction on $\hat{S}$ for a fair comparison with other automatic colorization works.

The advantages of our method are as follows: 1) Semantic Segmentation can be independently improved by leveraging previous research works related to semantic segmentation. 2) Our colorization network can learn the perfect features combining the human-annotated semantic map and gray values for colorization. Unlike traditional methods using concatenation for multiple inputs, our colorization network has two inference streams, and only gray features are normalized while semantic features still have the original range. 3) The user can constantly adjust the coarse semantic map from Semantic Segmentation to guide the colorization network.

\subsection{Semantic Segmentation Network}
Semantic information plays an important role in colorization. It can be associated with gray-scale values to provide plausible colors. Therefore, an accurate semantic map is crucial for our colorization network. Since it is inconvenient for the user to fulfill the map, we use Semantic Segmentation in our system to support our colorization and the user at the beginning. In this study, we adopt the GridNet \cite{fourure2017residual} for our Semantic Segmentation. The network is customized to have five rows and six columns with the channel dimensions of each row as $[16, 32, 64, 128, 256]$. All convolution modules use kernel size $3 \times 3$, padding of $1$, a stride of $1$, excluding sub-sampling convolution modules. The down-sampling layer uses a convolution module with a stride of $2$ to reduce the spatial size, while the up-sampling layer uses transposed convolution module with a stride of $2$. We apply cross-entropy segmentation loss $\mathcal{L}_{seg}$ to optimize the error between the estimated semantic map and its human-annotated map.

\subsection{Colorization Network}
Our colorization network generates the colors for a black-and-white photo based on its semantic map, representing an image-to-image translation task. We thus modify the well-known U-Net \cite{ronneberger2015u}, which achieves the outstanding performance in transforming an image to another image \cite{isola2017image}. For the more proficiency of our colorization, we modify the U-Net to have two streams, one for the gray-scale image and another one for the semantic map, as shown in Figure~\ref{fig:color}. Consequently, we can only normalize the gray features while the value range of semantic features is preserved. Our modified U-Net contains two parts: encoder and decoder. We use the same weights of the encoder for two data streams since both inputs have the same pose, retaining the characteristic of one-stream inference when inputs are concatenated.

Each part of the network has five layers consisting of two types: Convolution Layer (CL) and Deconvolution Layer (DL). CL is used in the encoder to reduce the spatial dimension by a convolution with a stride of $2$. Meanwhile, DL used in decoder expands the spatial dimension of features back to the original size by a transposed convolution with a stride of $2$. The extracted low-level features from the semantic information and gray image will be concatenated and transferred from encoder to decoder via skip connections. A layer has two convolution modules, and where each module is followed by a Leaky Rectified Linear Unit (Leaky ReLU) with a negative slope of $0.01$. The first convolution module in a layer is for sub-sampling, excluding the first and final layers. All convolution modules use a kernel size of $3 \times 3$. The depth dimensions in the order of inference flow are $[32, 64, 128, 256, 512]$ in the encoder part and will be doubled up to $[1024, 512, 256, 128, 64]$ in the decoder part because of the concatenation of extracted features from two streams of data, as shown in Figures \ref{fig:color} and \ref{fig:components}.

\textbf{Instance Normalization (IN).} Our network can provide plausible colors based on gray values and a semantic map. However, the performance of our colorization may be harmed by the complicated contrast information, as described in \cite{insnorm_ulyanov16} regarding image style transfer and in Figure \ref{fig:contrast} regarding image colorization. There still exists color bleeding, the improperly generated colors because of context confusion, even if an accurate semantic map is given. To address these problems, we utilize Instance Normalization (IN) to remove the instance-specific information of the gray features before the concatenation with semantic features.

\textbf{Loss function.} Inspired by \cite{zhang2017real}, we optimize the predicted colors $\hat{Y}$ and the ground truth $Y$ using Huber loss as:
\begin{equation}
 \mathcal{L}_{color}(Y, \hat{Y}) = \begin{cases}
\frac{1}{2}{( Y - \hat{Y}) }^{2}  & \quad \text{for } \vert Y - \hat{Y} \vert \leq \delta \\
\delta \vert Y - \hat{Y} \vert - \frac{1}{2} \delta^{2} & \quad \text{otherwise}
\end{cases}
\end{equation}

where $\delta=1$.

\section{Data Preparation and Training Details}
\textbf{Training dataset}. Our networks are trained on PASCAL-Context dataset \cite{mottaghi_cvpr14}, which has a semantic map completely annotated with varied categories. The dataset contains $10,103$ images with $59$ classes of the most frequent appearances, as considered in \cite{mottaghi_cvpr14}. Regarding the color space, we use the CIE \textit{Lab}, which allows us to separate gray-scale and color components from a typical colorful image. CIE \textit{Lab} consists of $L$ for the lightness and $a$, $b$ for the color components. The values of $L, a, b$ are scaled and normalized into the range $[-1,1]$.

\textbf{Data augmentation.} To diversify our training data, we scale the training images from varied sizes to $360 \times 360$, randomly crop them to $352 \times 352$, and randomly flip in horizontal and vertical ways.

\textbf{Training details.} We train our models with Adam optimizer \cite{kingma2014adam} with $\beta_{1} = 0.9$, $\beta_{2} = 0.999$, initial learning rate of $0.0001$, batch size of $16$. Most of the models are trained for $1000$ epochs on Tesla V100 in one week.


\section{Experiments}
\label{sec_perform}
This section demonstrates our effectiveness in providing plausible colors and handling colorization-related problems such as monochrome, color bleeding, semantic detection faults, etc., even on an old photo or a painting. Furthermore, we compare our colorization (automatic) with recent automatic colorization works \cite{zhang2016colorful, larsson2016learning, iizuka2016let, zhang2017real, Su-CVPR-2020}. Besides, an interactive application is presented for adjusting semantic faults with a user-friendly interface.

\begin{figure}[ht]
\centering
\includegraphics[width=\textwidth]{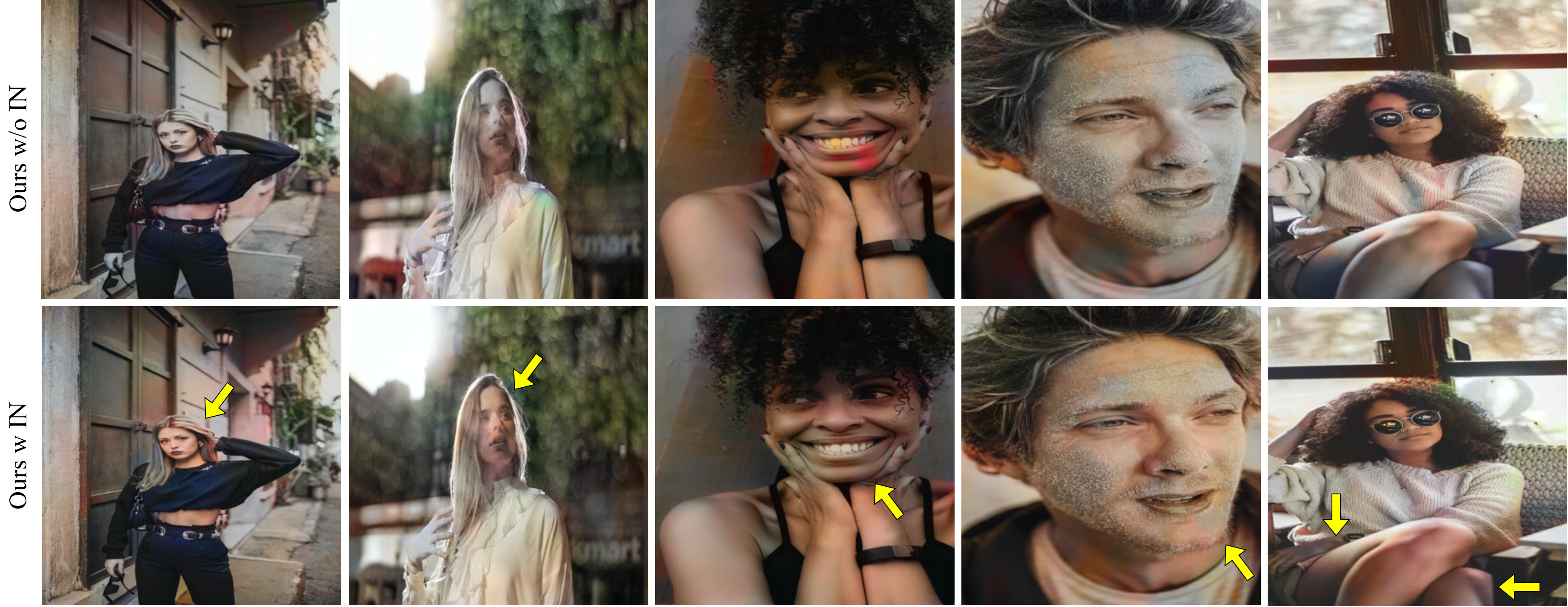}
\caption{The effectiveness of Instance normalization (IN) in simplifying gray features and removing instance-specific information. Artifacts are removed to provide harmony and plausibility in color. \textit{Top-bottom}: Generated colors by our network trained without and with IN.}
\label{fig:insnorm}
\end{figure}

\subsection{Ablation Study on Instance Normalization (IN)}
The contrast of black-and-white input image significantly affects the generated colors, and worsens colorization performance when the contrast becomes low, as proved in Figure \ref{fig:contrast}. Therefore, IN \cite{insnorm_ulyanov16} is also a missing ingredient for image colorization and the gray features should be normalized by IN. However, the typical inference style of U-Net needs all inputs to be concatenated, forcing the semantic features to be also normalized. Hence, we create two streams of data and extract the gray feature maps from the black-and-white input $X$ and the semantic feature maps from the semantic map $S$ separately. By doing so, we can simplify the gray features while keeping the range of semantic values intact before transferring the concatenated features to the decoder. To prove the performance of simplifying the gray feature maps, we train two models with/without IN in the same condition and compare them qualitatively. As a result, the model with IN gives more harmonious and plausible colors than the model without IN, as shown in Figure \ref{fig:insnorm}.

\begin{figure}[t]
\begin{center}
\includegraphics[width=\textwidth]{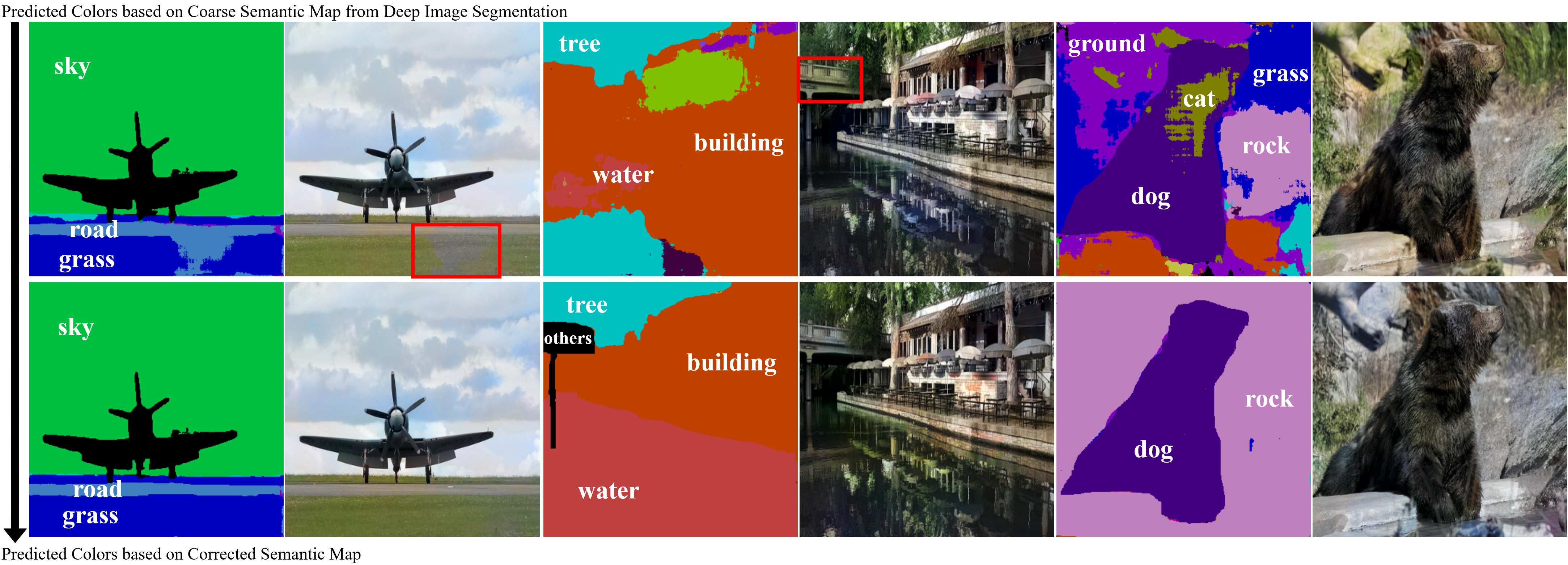}
\end{center}
\caption{Color/Semantic Correction. \textit{From left to right}, we show three samples with the corresponding problems: (a) patching the semantic fault on the region of grass, (b) fixing inharmonious color and green color bleeding on the bridge, (c) fixing the implausible green color on the bear and rock. \textit{Top row} shows the results of our automatic colorization before the adjustment, \textit{bottom row} shows the results after the adjustment. We highlight several parts for a quick comparison.}
\label{fig:correction}
\end{figure}

\subsection{Color/Semantic Correction}
The coarse semantic map from Semantic Segmentation usually causes color inconsistency. By simulating the human-like action in coloring, our colorization system allows users to correct the semantic faults of the segmentation model by using strokes. As a result, by correcting the semantic map, the color bleeding on the "\textit{grass}", "\textit{bridge}", and "\textit{bear}" is eliminated, as shown Figure \ref{fig:correction}.

\begin{figure}[t]
    \centering
    \includegraphics[width=0.8\textwidth]{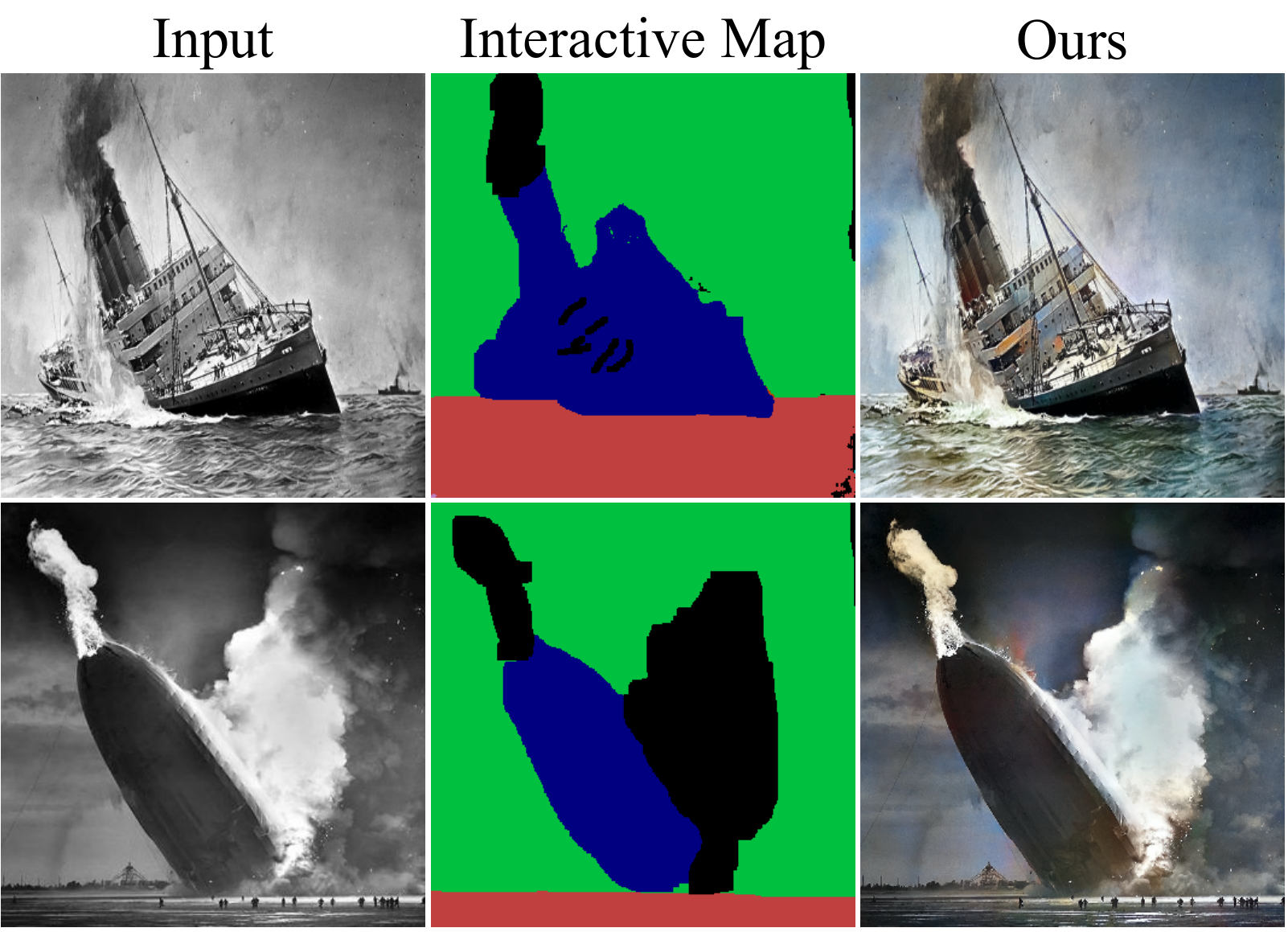}
    \caption{Our performance on the old photo of the Hindenburg Disaster on May 6, 1937, and the painting of Sinking of the RMS Lusitania on May 7, 1915.}
    \label{fig:old}
\end{figure}

\subsection{Our performance on the old photo and painting.} Bringing colors to old black-and-white photos is a meaningful goal that most colorization works aim to. However, the model is trained on high-quality photos taken by modern cameras from COCO dataset \cite{caesar2018coco}, old photos and paintings are thus out of our training distribution. Even so, this work still provides plausible colors for the old photo of the Hindenburg Disaster on May 6, 1937, and the painting of Sinking of the RMS Lusitania on May 7, 1915, as shown in Figure \ref{fig:old}.

\begin{figure}
    \centering
    \includegraphics[width=\textwidth]{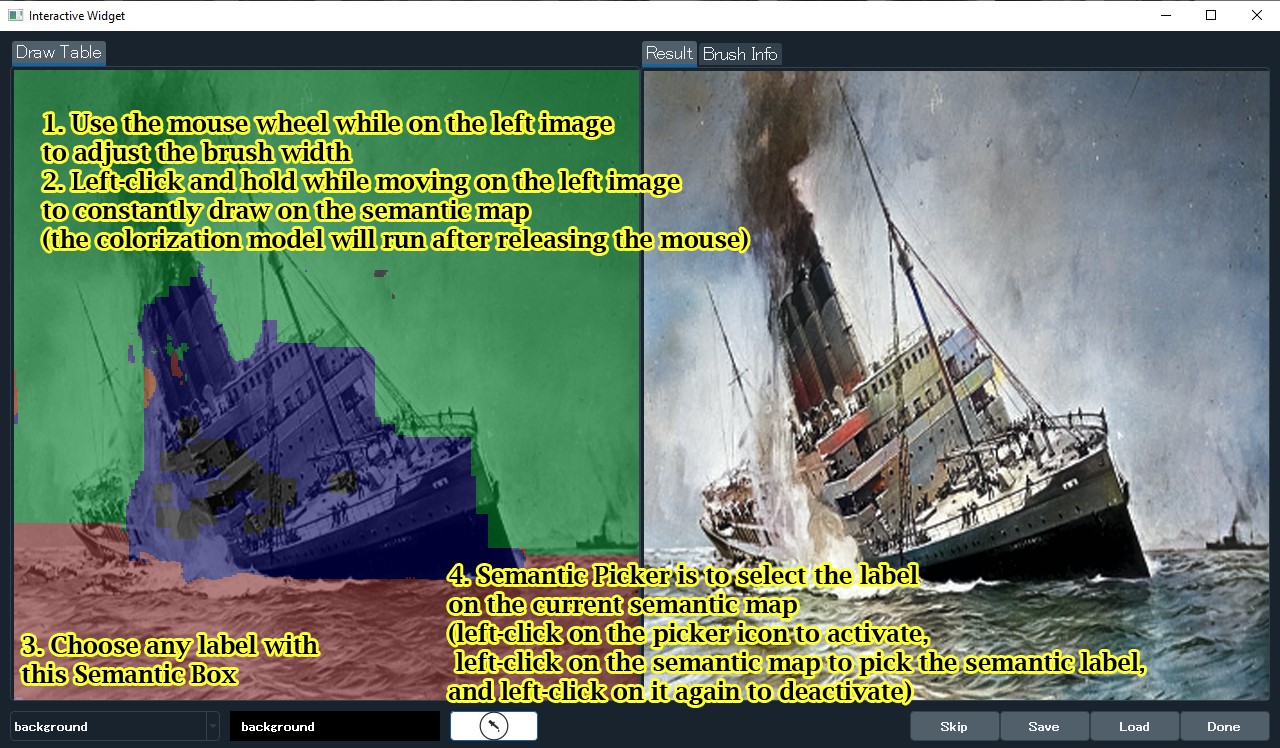}
    \caption{Our interactive application for semantic information adjustment with guidance.}
    \label{fig:app}
\end{figure}

\subsection{Interactive Application}
Generated colors are implausible due to the improper semantic values given by the segmentation model or the colorization model's faults. To understand our learned models, we build an interactive application to remove/change semantic information, as described in Figure \ref{fig:app}. Besides, our application achieves a real-time performance with the computation time of coloring a $352 \times 352$ image of \textbf{8} milliseconds on VGA NVIDIA GeForce GT 730.

\begin{figure}[t]
\begin{center}
\includegraphics[width=\textwidth]{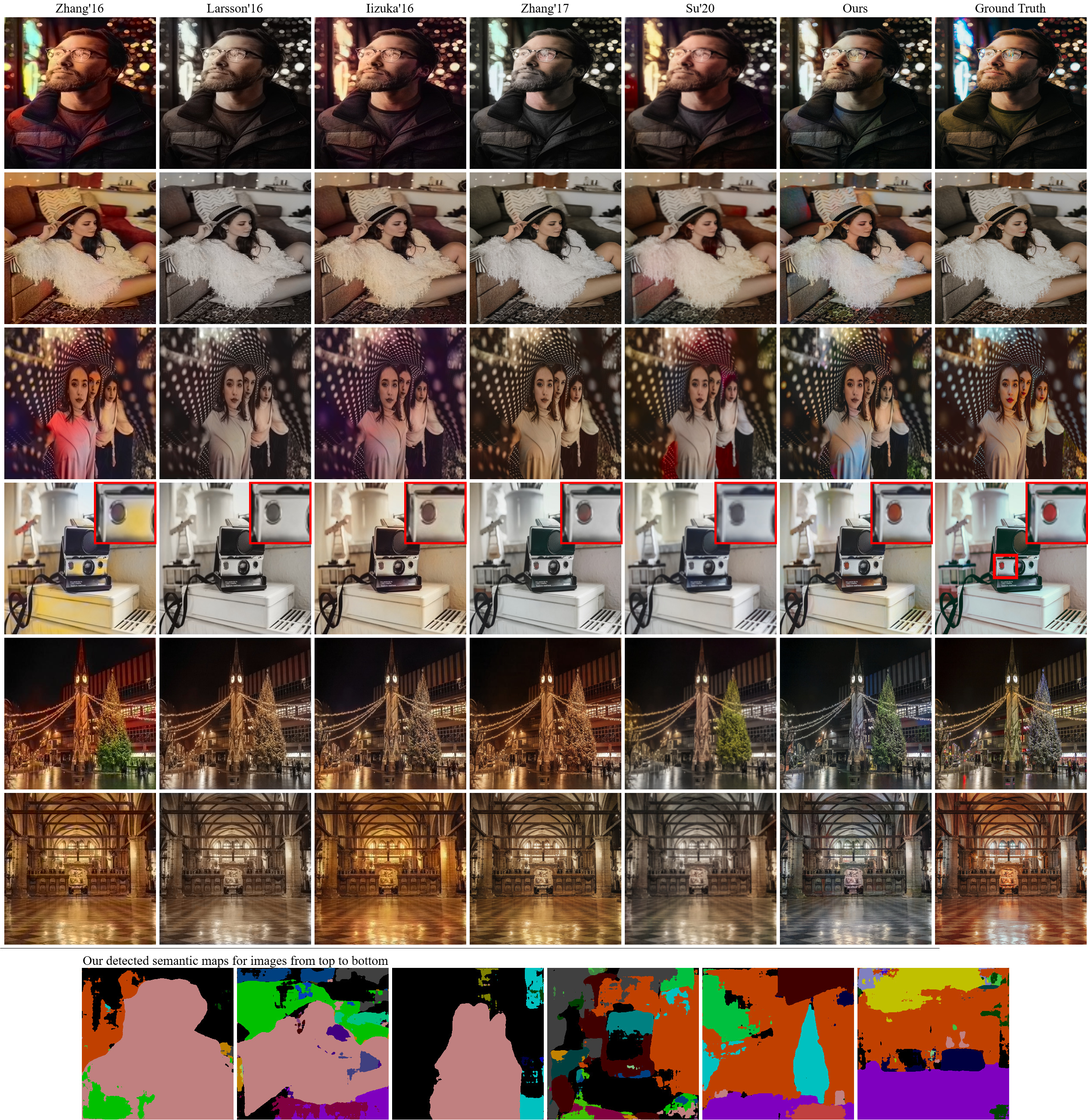}
\end{center}
\caption{Comparison in automatic colorization. \textit{Top part}, \textit{Left to right}: Zhang'16 \cite{zhang2016colorful}, Larsson'16 \cite{larsson2016learning}, Iizuka'16 \cite{iizuka2016let}, Zhang'17 \cite{zhang2017real}, Su'20 \cite{Su-CVPR-2020} and our semantic-driven work. Six compared images are shown in \textit{each row}. \textit{The final row} shows the coarse semantic maps we use for our colorization.}
\label{fig:previous}
\end{figure}

\subsection{Comparison with Previous Works in Automatic Colorization}
We compare our work with the recent automatic colorization works of Zhang et al. \cite{zhang2016colorful} (\textit{Zhang'16}), Larsson et al. \cite{larsson2016learning} (\textit{Larsson'16}), Iizuka et al. \cite{iizuka2016let} (\textit{Iizuka'16}), and Su et al.\cite{Su-CVPR-2020} (\textit{Su'20}), and the interactive colorization of Zhang et al. \cite{zhang2017real} (\textit{Zhang'17}) without guided colors. Additionally, inspired by \cite{zhao2018pixel}, we also conduct a user study on user preference in Semantic Correctness, Saturability, and Edges Keeping. 

\textbf{Qualitative Comparison.} In the domain of automatic colorization, \textit{Zhang'17} without user suggestions is guided by a black image and a zero mask. Meanwhile, our method uses the coarse semantic map detected by our re-trained Semantic Segmentation. In a comparison with \textit{Zhang'16}, \textit{Larsson'16}, \textit{Iizuka'16}, \textit{Zhang'17}, and \textit{Su'20}, our semantic-driven work can provide the more plausible colors for specific objects without facing the monochrome problem, as shown in shown in Figure \ref{fig:previous}. Particularly, "\textit{the guy's face}" in our result has a plausible color with a distinct tone from \textit{his coat} in \textit{row 1}; meanwhile, other works give the same tone. Plus, our work provides a better \textit{skin color} for \textit{the girl} with the more distinguished tone from \textit{the background} than other works in \textit{row 2}. Regarding row 3, our work and Su'20 can generate a distinct color for \textit{the girls} from \textit{the background} while other works suffer from the monochrome problem. However, the color of Su'20 is polluted by abnormal red color, reducing the naturalness of the photo. The next three rows also show our effectiveness in coloring small details and specific objects with plausible and varied colors compared with other works, even though our coarse semantic maps do not have any user intervention and cause a few artifacts. In summary, the colors generated by our semantic-driven colorization are able to outperform the colors by typical and recent research works qualitatively.

\begin{table*}[t]
\centering
\begin{tabular}{|l|c|c|c|c|c|}
\hline
& \multicolumn{1}{l|}{Semantic Correctness} & \multicolumn{1}{l|}{Saturability} & \multicolumn{1}{l|}{Edges Keeping} & \multicolumn{1}{l|}{Naturalness} & \multicolumn{1}{l|}{Naturalness (*)} \\ \hline
Original & 85.76 & 87.88 & 89.82 & 87.8 & 66.62 \\ \hline
Iizuka'16 \cite{iizuka2016let} & 67.16 & 65.5 & 70.52 & 67.7 & 43.5 \\ \hline
Zhang'16 \cite{zhang2016colorful} & 61.02 & 63.88 & 62.96 & 62.6 & 42.0 \\ \hline
Larsson'16 \cite{larsson2016learning} & 67.3 & 62.62 & 70.26 & 66.7 & 45.13 \\ \hline
Zhang'17 \cite{zhang2017real} & 64.8 & 60 & 70.98 & 65.3 & 43.0 \\ \hline
Ours & \textbf{71.9} & \textbf{71.62} & \textbf{71.1}  & \textbf{71.5} & \textbf{59.63} \\ \hline
\end{tabular}
\caption{A user study on 3 criteria such as Semantic Correctness, Saturability, and Edges Keeping with the scale of 100. The result shows that our generated colors based on semantic information are highly competitive to the typical colorization works. A higher score means better performance.}
\label{tab_user1}
\end{table*}

\textbf{User Study.} Our target is to generate the naturalness of color that can fool people. Therefore, inspired by \cite{zhao2018pixel}, we briefly conduct a user study on the correctness of colors, saturation, and color overflow. Concretely, the participant will score the ground-truth images and results from previous works \cite{iizuka2016let, zhang2016colorful, larsson2016learning, zhang2017real} shown one-by-one based on three criteria such as \textbf{Semantic Correctness}, \textbf{Saturability}, \textbf{Edges Keeping} with a scoring scale of $0-100$, where $0$ means a machine definitely generates the photo, and $100$ means the photo is original. Naturalness is the average score of three criteria. Additionally, we compare all colorization works in Naturalness (*) by showing the ground-truth image and results from different methods together (positions are shuffled), then let the participants score them. Thus, the performance of each method will be sorted by participants. As shown in Table \ref{tab_user1}, our work outperforms the previous works in fooling humans on three mentioned criteria with a Naturalness of \textbf{71.5\%}. Besides, when all images are shown together, our results have the highest probability of being real as \textbf{59.63\%} compared with the previous works; meanwhile, participants think there is a probability of \textbf{66.62\%} that the ground-truth images are original.

\section{Discussion}
\label{sec_discuss}

\begin{figure}[t]
    \centering
    \includegraphics[width=0.8\textwidth]{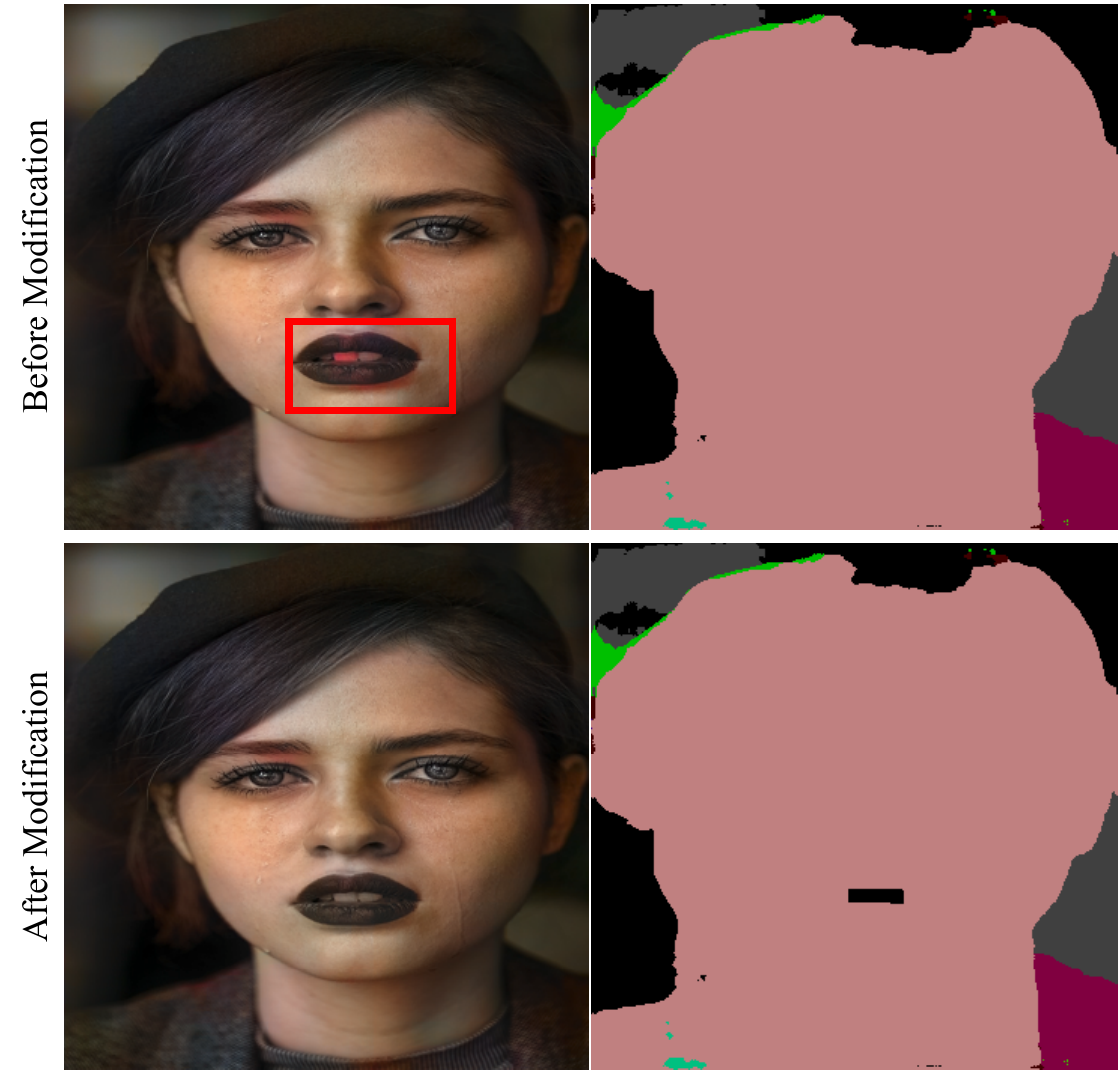}
    \caption{Fixing implausible colors by unreasonably modifying semantic information. We eliminate the implausible red color on her teeth by changing the label of semantic values on her teeth from "person" to "background".}
    \label{fig:unreason}
\end{figure}

\textbf{An implausible color occurs over a region with correct semantic values.} We make the generated colors more plausible by establishing a constraint between the grayscale values and semantic information; however, our colorization model is still data-driven. Therefore, an implausible color can still occur, even when semantic values are correct. For example, there is an \textit{implausible red color}, which should have been for \textit{lips}, on \textit{the girl's teeth}. It may be due to our colorization model trained mostly on the label "person" with a closed mouth. To solve this issue, we change the semantic values of the region where implausible colors occur to correct the color, as shown in Figure \ref{fig:unreason}.

\textbf{Multi-color Suggestion and User-guided Colorization.} Currently, our colorization system directly predicts the natural color for specific objects based on their semantic information. To enhance colorization performance, we can leverage more types of user guidance such as guided color dots \cite{zhang2017real} and edges \cite{kim2021deep}. Thus, the suggested color is more natural, and the semantic boundaries are also improved. Also, the coarse semantic map can be improved automatically via user suggestions.

\textbf{Coarse Semantic Map in Automatic Colorization.} 
In fact, the deep neural network can make mistakes as Zhang et al. \cite{zhang2017real} mentioned about the generated color, Larsson et al. \cite{larsson2016learning} mentioned about the unrecognizable objects, and Xiao et al. \cite{xiao2018interactive} mentioned about the inaccurate semantic segmentation. That causes many problems in colorization, such as color inconsistency, color bleeding, etc. Our target is to solve the problems using suitable semantic information. However, our method also makes mistakes in segmentation and colorization. Consequently, our results also have some symptoms of the mentioned problems, such as color bleeding, color inconsistency. Thanks to IN efficiency (that we present in Section~\ref{sec_perform}), those problems can be significantly addressed, but not entirely. Therefore, we build an application to adjust the visible semantic map in the middle of our colorization framework.

\section{Conclusion}
As a human experience in black-and-white photo coloring, humans step-by-step recognize the objects in the photo, imagine the plausible colors for specific objects as we have seen them many times in real life (semantic-driven), and finally colorize it. In this work, we simulate that human-like action and let our network understand the black-and-white photo and then colorize it. Therefore, the predicted semantic information of the whole colorization framework is now visible and can be adjusted with our interactive application. Additionally, we also prove that Instance Normalization \cite{insnorm_ulyanov16} is also a missing ingredient for colorization, then re-design the inference flow of U-Net to have two streams of data, providing an appropriate way of normalizing the feature maps from the black-and-white image and its semantic map. As a result, our method can generate plausible and varied colors with more naturalness than previous works for specific objects. Besides, our generated colors between objects are distinct, suppressing the monochrome problem in colorization.

\bibliographystyle{unsrt}
\bibliography{citations}

\end{document}